\documentclass[12pt,journal]{article}

 \usepackage{times}  %Required
\usepackage{helvet}  %Required
\usepackage{courier}  %Required
\usepackage{url}  %Required
\usepackage{graphicx}  %Required
\usepackage{caption}
\usepackage{subcaption}
\usepackage{wrapfig}
\usepackage{tabularx}
\usepackage{placeins}
\usepackage{dblfloatfix}
\usepackage[table]{xcolor}
\usepackage{multirow}
\usepackage{algorithm}
\usepackage{algpseudocode}
\usepackage{pifont}

\usepackage{amsmath}
\usepackage{amssymb}
\usepackage{amsfonts}
\usepackage{amsthm}
\usepackage{mathtools}
\usepackage{bm}
\usepackage{mathrsfs}
\usepackage{color, colortbl}
\usepackage{verbatim}

\usepackage[hang,flushmargin]{footmisc}   % flush footnotes

\definecolor{LightCyan}{rgb}{0.88,1,1}
\definecolor{LightGreen}{rgb}{.88,1.,0.88}

\def\w{\mathbf{w}}

\def\w{\mathbf{w}}

\definecolor{mycolor}{cmyk}{0.0,0.2,0.9,0.0}

\begin{document}
\title{Discovering Molecular Functional Groups Using Graph Convolutional Neural Networks}

\author{Philip~E.~Pope*, Soheil~Kolouri*,
         Mohammad Rostami*\footnote{These authors contributed equally.},\\ Charles E. Martin,
        and~Heiko~Hoffmann }

\maketitle

%\IEEEtitleabstractindextext{%
\begin{abstract}
Functional groups (FGs) are molecular substructures that are served as a foundation for analyzing and predicting chemical properties of   molecules. Automatic discovery of FGs will impact various fields of research, including medicinal chemistry and material sciences, by reducing the amount of lab experiments required for discovery or synthesis of new molecules. In this paper, we investigate methods based on graph convolutional neural networks (GCNNs) for localizing FGs that contribute to specific chemical properties of interest. In our framework, molecules are modeled as undirected relational graphs with atoms as nodes and bonds as edges. Using this relational graph structure, we trained GCNNs in a supervised way on experimentally-validated molecular training sets to predict  specific chemical properties, e.g., toxicity. Upon learning a GCNN, we analyzed its activation patterns to automatically identify FGs using four different explainability methods that we have developed: gradient-based saliency maps, Class Activation Mapping (CAM), gradient-weighted CAM (Grad-CAM), and Excitation Back-Propagation. Although these methods are originally derived for convolutional neural networks (CNNs), we adapt them to develop the corresponding suitable versions for GCNNs. We evaluated the contrastive power of these methods with respect to the specificity of the identified molecular substructures and their relevance for chemical functions. Grad-CAM had the highest contrastive power and generated qualitatively the best FGs. This work paves the way for automatic analysis and design of new molecules.
\end{abstract}

% \begin{IEEEkeywords}
% Graph convolutional neural networks, Molecular functional groups, Neural network explainability methods.
% \end{IEEEkeywords}}


%\IEEEdisplaynontitleabstractindextext

% \fi
%\IEEEpeerreviewmaketitle

%\IEEEraisesectionheading{\section{Introduction}\label{sec:introduction}}
\section{Introduction}\label{sec:introduction}
Despite limited number of chemical elements, these elements combine in considerably different arrangements  and create a far greater number of molecules which have a wide range properties. Given a particular application, it is crucial to use  suitable materials with specific desired properties to reach the end goal. Suitable materials may be discovered in nature or designed and synthesized by human experts. Classic molecule discovery methods in chemistry and material sciences are based on   semi-trial/error procedures that are economically expensive and time consuming as the hypothesis classes, i.e. chemical space of molecules, are usually huge for even specific classes of molecules. For these reasons, automatic computer-aided  design of molecules with desired properties is an emerging field with important applications including drug discovery~\cite{gawehn2016deep}, molecular cancer diagnosis~\cite{sevakula2018transfer}, and material design~\cite{gomez2018automatic}.

 Computer-aided  design can help to minimize the  hypothesis classes and speed up the experimental procedures by eliminating uninformative experiments.  Moreover,  once a new class of molecules is discovered, it is important to investigate the commonalities between these molecules, i.e. identifying patterns of a particular class.
Despite the complex structure of many molecules, in particular organic molecules, specific properties, e.g., toxicity or solubility in water, may be caused by specific  substructures of atoms within a molecule. These substructures are called functional groups (FGs); e.g., toxophores are a specific type of FG's that produce toxicity in toxin molecules~\cite{ridings1996computer}. FGs impose specific properties on molecules because they can participate in specific chemical reactions that give rise to a specific property. The study of common FGs is central to organic and inorganic chemistry~\cite{critchfield1963organic},   e.g., the hydroxyl FG ($-OH$), as they can be used to systematically predict behavior of a compound in chemical reactions. Identification of FGs can hint how to design molecules that posses or lack particular properties. Our goal in this paper is to propose a new method to identify FGs.

Similar to the experiment-based molecular design, the experimental approach to FG discovery requires high-precision instruments to collect data using either  Fourier Transform
Infrared spectroscopy (FTIR)~\cite{kong2007fourier} or Mass spectroscopy (MS)~\cite{domon2006mass} for each individual molecule, followed by many hours of expert supervision to analyze the data for a class of molecules.
There are already about 50 millions of discovered molecules and experimental search in even a  limited subset of this hypothesis space can be time-consuming. Additionally,  many more potentially stable molecular structures are yet to be discovered, requiring an additional step of synthesis to perform the above tests. In modern applications such as drug discovery~\cite{gawehn2016deep}, the discrete hypothesis space is estimated to have a size of $10^{23}$ to $10^{60}$ molecules~\cite{polishchuk2013estimation} and even small changes in the molecular structure can change properties in a discrete space dramatically. This makes  searching the space experimentally  infeasible.
Fortunately, recent efforts in combining computational chemistry and machine learning (ML) leverage data driven methods suggests possibility of efficiently narrowing down this search space~\cite{duvenaud2015convolutional,gomez2018automatic,wu2018moleculenet,butler2018machine}. This stems from success of machine learning in areas such as computer vision and natural language processing, were predictive models can help to investigate unexplored instances of data via successful generalization of past experiences by identifying  patterns.

Recent success of ML with human-level performance in some  computer vision applications is due to reemergence of deep  neural networks~\cite{krizhevsky2012imagenet}. Deep nets have easened the tedious  engineering task of feature extraction from data because they are trained in an end-to-end data-driven scheme which automatically extract suitable features for a given task from data. Using the specific class of Deep convolutional neural networks  (CNNs) on vision tasks  has led to human-level performance on classification~\cite{huang2017densely} and object detection~\cite{redmon2017yolo9000}  tasks.  The success of CNNs may be attributed to convolutional layers, which regularize the network by reducing the number of learnable parameters and allow training deeper networks for multi-level abstraction of feature extraction. Since the neurons in CNN structure receive input from a number of neighboring nodes, they specialize to attend specific regions of input. As a result, the hierarchical features may be used to localize signal to regions of the input, giving a means of interpreting and explaining decisions by different layers of the network~\cite{zhou2016learning}.
Despite this success, traditional deep CNNs are designed for Euclidean space where data is defined on a structured  grid, e.g. domain of natural images. This is because convolution is an operation defined on Euclidean space for rectangular lattices. For this reason, CNNs cannot be directly used on  domains with other data structures such as graph-structured data.  %

A recent variant of CNNs designed for graph-structured data are graph convolution neural networks (GCNNs)~\cite{duvenaud2015convolutional,gilmer2017neural,schlichtkrull2018modeling}. GCNNs are built upon  generalizing the definition of convolution to non-Euclidean data structures that can be modeled by graphs. By adopting graph-specific versions of common types of CNN layers such as convolutional, max-pool, and batch-normalization layers, multi-layer deep GCNNs can be formed similar to CNNs.  As a result, GCNNs inherit properties like shared weights that regularize the network according to the relations between the nodes that that are captured by the edges. Similarly, deep hierarchical feature distillation emerge in GCNNs which have led to promising results in classifying graph-structured data, including knowledge graphs and social networks~\cite{kipf2016semi,schlichtkrull2018modeling}.
Building upon the success of CNNs in computer vision, a recent line of research has applied GCNNs to  atomic and molecular applications~\cite{duvenaud2015convolutional,kearnes2016molecular,schutt2017schnet,wu2018moleculenet,gomez2018automatic}. In these methods, molecules are modeled as graphs, where the graph nodes represent the atoms, and the graph edges (potentially weighted) represent the chemical bonds and their types, and learning is performed on the molecule-level.  Inheriting  properties of CNNs, GCNNs have lead to promising results in this area as the properties of molecules stems from the particular arrangement of the forming atoms.

In this paper, we  propose explainability  methods for GCNN to determine localized parts of a graph which correspond to a specific classification decision, as inspired by related work on images~\cite{zhou2016learning}. We employ our idea\footnote{Partial early results of this paper are presented as an oral presentation at CVPR 2019~\cite{pope2019explainability}} on molecular classification tasks and demonstrate that our approach can be used for identifying FGs. We   adapt and extend existing explainability  methods for CNNs to become applicable on GCNNs. We propose based on four different explainability methods: gradient-based saliency maps, Class Activation Mapping (CAM), Gradient-weighted Class Activation Mapping (Grad-CAM), and Excitation Back-Propagation (EBP). We evaluate the performance of these methods with respect to contrastiveness (class-specific localization) and sparsity of localization and qualitatively compare the localization heatmaps over the graph structure. We then use our methods to investigate explainability of GCNNs on  molecular datasets. Our experiments confirm that highlighted structural components can show known FGs that correspond to a certain chemical property, leading to possibility of discovering new FGs.

The rest of the paper is as follows. In Section~\ref{sec:UDArelated1}, we review recent related works on identifying the FGs. In Section~\ref{sec:UDArelated2}, we survey  the explainability methods  for CNNs that we extend in our paper. We explain that how the CNN explainability methods can be adapted to be used for GCNNs in Section~\ref{sec:UDArelated3}. Section~\ref{sec:UDArelated4} is devoted to experimental validation of our approach on two molecule datasets to demonstrate that our approach can identify FGs. Finally, the paper is concluded in Section~\ref{sec:UDArelated5}.

 \section{Related Work}
 \label{sec:UDArelated1}
Computer-aided molecular design using neural networks is not a recent idea in chemistry~\cite{gasteiger1993neural,devillers1996neural,schneider1998artificial}, where the initial idea was to use neural networks for prediction tasks in chemistry, e.g., predicting solubility level of different materials in a solvent, but later possibility of discovering new molecules.   Following the breakthrough of deep learning, various modern deep network structures have also been used in chemistry and biochemistry including   CNNs~\cite{wallach2015atomnet}. This has led to the state of the art performance prediction tasks  in applications such as drug discovery~\cite{gawehn2016deep} and predicting chemical properties~\cite{lusci2013deep}.
More recent structures such as Generative Adversarial Networks (GANs)~\cite{kadurin2017drugan} and Variational Autoencoders (VAEs)~\cite{liu2018constrained} have been used as generative models to generate potential novel molecules with desired properties which dramatically can improve synthesis tasks.
 The major challenge of employing ML techniques on molecular level chemistry and biochemistry is data representation. Most neural network structures are designed to  receive and process multidimensional arrays such as images as their input. For this reason, various approaches have been   developed in the literature to tackle this challenge for graph-structured data.  To overcome this challenge, we either need   to convert  molecules in a dataset to fixed-size arrays to use the existing networks or adapt and enable the network structures to receive graphs directly at their input.

 When a molecule is converted into a vector, the resulting vector must encode the important structural information of the corresponding molecule. In the ideal case, the representation should be unique and invertible to guarantee lossless representation, but in practice most vector representation methods for molecules are only invertible.
 A simple data representation method is to convert molecules to binary vectors of fingerprints vector based on presence or absence of a property~\cite{mayr2016deeptox,kadurin2017cornucopia,sanchez2018inverse}. A more common approach is to  process and parse   Simplified Molecular-Input  Line-Entry  System  (SMILES) representation of molecules which represents molecules as strings of text~\cite{kadurin2017drugan}. The text embedding approaches or simply one-hot vector conversion then can be used to convert the SMILES string into a vector. Many ML models have been adopted in applications involving molecules using SMILES representation~\cite{duvenaud2015convolutional,segler2017generating,gilmer2017neural,elton2019deep}.
 Note however, since SMILES representation is not unique for a given molecule and are drawn with specific rules, the representing vectors may capture rules of building SMILES strings rather the  structural information about  the underlying molecules. Additionally, since a single character perturbation in SMILES representation of a molecule can change the underlying molecule significantly, learning from SMILES strings is challenging due to sensitivity of the representation~\cite{liu2017retrosynthetic}.

A major benefit of vector representation is that we can then  employ most standard ML models that exist for predicting chemical characteristics of molecules~\cite{dahl2014multi}. However, similar to classic AI research areas such as computer vision, finding the proper feature extraction method which works well for a given application, can be difficult and restrictive.
 Proper feature extraction has   always been a major challenge for ML but the recent reemergence of deep neural architectures, including CNNs, has led  to automation of the process of feature extraction in an end-to-end  data driven scheme based on suitability for a particular application.

 Automatic feature extraction is the major reason behind the   success and popularity of deep learning in computer vision applications. Additionally, CNNs have been used for data generation which enables a user to generate synthetic samples for a given class~\cite{goodfellow2014generative}. This means that a user can generate samples that possess a predetermined property  which makes them suitable for applications involving discovery. The challenge of applying CNNs on molecule datasets is that CNNs can only receive   data  that is structured as a rectangular lattice
 as their input, e.g., images. This limitation has been circumvented by the invention of graph convolutional neural networks (GCNN). GCNNs adapt and change structure of CNNs to make them applicable on non-Euclidean spaces of graph-structured data such as molecules.
 As a result, data representation challenge is resolved directly by changing the model structure.
 Similar to CNNs, GCNNs are able to learn descriptive features automatically that outperform engineered features, enabling
 GCNNs to achieve state-of-the-art performance on several chemical prediction tasks, including toxicity prediction~\cite{kearnes2016molecular}, solubility~\cite{duvenaud2015convolutional}, and energy prediction~\cite{schutt2017schnet}. In this work,  our goal  is to move one step beyond the prior works in the literature which focus on prediction tasks within chemistry and biochemistry. We develop explainability methods for GCNNs to investigate decision process by these networks for identifying potential FGs in a class of molecules.

Deep nets are   considered to be  black boxes to a large extend. In other words, although   deep nets perform well on many tasks, we do not have clear understanding about the reason behind their good performance. However, simply performing well without understanding deep models is not sufficient for further progress. Unlike some existing ML models which are based on logical and symbolic reasoning, interpreting the data processing procedure by deep nets is quite challenging.
The reason is that deep nets have a huge number of learnable parameters and hyper-parameters and are highly non-linear and non-convex models. Moreover, the corresponding empirical risk minimization optimization problems are non-convex with non-unique solutions. Additionally, several different stochastic optimization methods can be used for training deep nets. As a result, it is not that straightforward to determine the decision boundaries or contribution of particular learnable parameters for a given trained deep network.
 A good explainability method can help us to improve the performance of existing deep nets because explainability methods help to discover biases and weaknesses of deep network models. Moreover, these methods also could be used as an explanatory tool to ensure humans that a particular network attends to intuitively sensible areas of the input to enable humans to trust black-boxes. Beyond a pathway to trust deep nets to replace humans, these methods can   help human experts to learn from deep nets because as   some decisions by deep network models might be completely new for humans, yet quite informative~\cite{silver2016mastering}.  For these reasons, developing explainability methods for GCNNs can be helpful beyond our application of interest, i.e., discovering FGs in a class of molecules.

Our approach is to benefit from the existing explainability methods for CNNs to develop explainability methods for GCNNs.
Several explainability methods have been devised for deep networks and specifically CNNs~\cite{simonyan2013deep,zhou2016learning,selvaraju2017grad,zhang2016top}. These methods enable one to probe a CNN and identify the important areas of the input data (as deemed by the network) that contributed to the network decision. We can use these methods to analyze the data processing procedure by a network and  explaining data representation  inside the network for a particular input.
Investigating CNNs using these methods indicate that predictive ability is not the only reason for superiority of CNNs. Similar to humans, they can identify important regions within an input that are important for prediction.
For example, in the area of medical imaging, in addition to classifying images having malignant lesions, they can be localized, as the CNN can provide reasoning for classifying an input image. In our work, we are interested in measuring the potential of these methods for discovery of FGs in  molecules as counterpart of important regions in images. This process can be particularly helpful for discovering FGs because, as opposed to images, humans cannot intuitively determine the relevant context within a molecule for a particular property of that molecule and only extensive experiments can help with this goal. In other words, deep nets can teach us what components in the input data points contribute to a specific property.

 The most straightforward approach for explaining data processing procedure in a deep network is generating a sensitivity map over the input data to discover the importance of the under lying substructures is to calculate a gradient map within a layer by considering the norm of the gradient vector with respect to an input for each network weight~\cite{simonyan2013deep}. As a result, we can identify areas of the input that cause high activation in the network and areas in the input that changes can affect the network decision.  However, gradient maps are known to be noisy and smoothening these maps might be necessary~\cite{smilkov2017smoothgrad}. More advanced techniques include Class Activation Mapping (CAM)~\cite{zhou2016learning}, Gradient-weighted Class Activation Mapping (Grad-CAM)~\cite{selvaraju2017grad}, and Excitation Back-Propagation (EB)~\cite{zhang2016top} techniques that improve gradient maps by taking into account some notion of context. These techniques have been shown to be effective on CNNs and can identify highly abstract notions in  images that help to solve a task, i.e., predicting the correct label in classification tasks.  Inspired by the explainability power of deep CNNs, our goal is to adapt these techniques and develop new versions for deep GCNNs to automatize discovery of chemical FGs for a particular behavior.

Our specific contributions in this paper are:
\begin{itemize}
\item Adapting explanation tools for CNNs to GCNNs with the application of discovering FGs for organic molecules, which can be potentially used for other graph-structured data.
\item Comparing the contrastive power and class specificity of the explainability methods in identifying FGs. Note that each of this properties can be important given the specific application.
\item Analyzing three molecular datasets and their identified FGs and to check whether our results can be validated by existing experimental results.
\end{itemize}
We envision that our proposed framework could help chemists with identifying new FGs, or at least suggesting potential choices, that have not been discovered before, reducing the search space and subsequently  experimental cost and the required time needed for this purpose.

\section{Explainability Methods for Convolutional Neural Networks}
\label{sec:UDArelated2}
Early explanations for CNNs were based on   interpreting the data representation inside the network pathway. Since CNNs are inspired from the nervous system structure, the   argument is based on the intuition that a deep net extracts hierarchical abstract features from the input data points, e.g., edges, colors, and shapes for images.
While this explanation may help to have a better intuition about operation of deep networks, it is not very helpful to explain decision procedure by a particular network.
Interpretable explainability methods focus on investigating the processing of individual data points and the reason behind decisions made by a network. For the case of CNNs, explainability methods investigate the network spatial attention on specific areas of the input images. By comparing the network attention on a number of similar data points, we can understand whether the decision process by a network is sensible, e.g., a network may attend to tires and windshields to deduce that an input object is a car.

Many explainability methods have been developed for CNNs recently. In this work, we focus on using four popular methods for CNNs: gradient-based saliency maps~\cite{simonyan2013deep}, Class Activation Mapping (CAM)~\cite{zhou2016learning}, Gradient-weighted Class Activation Mapping (Grad-CAM)~\cite{selvaraju2017grad}, and Excitation Back-Propagation (EBP)~\cite{zhang2016top}.  We adapt these methods to make them applicable to Graph Convolutional Neural Networks (GCNNs) and then compare and contrast the explanations that these methods generate as FGs on three molecular datasets.  We explore the benefits of a number of enhancements to these approaches.

A pioneer and probably the most straight-forward (and well-established) approach is to generate gradient-based saliency maps~\cite{simonyan2013deep}. The idea is to measure sensitivity of the network predictions given changes in the input. Intuitively, the spatial areas on the input that the network is sensitive about, play an important role on the network prediction. To generate a saliency map, one can simply differentiate the output of the model with respect to the model input, using automatic differentiation tool. A heat-map then can be created by using the norm of the gradient over input variables, indicating their relative importance. Note that the resulting gradient in the input space points in the direction corresponding to the maximum positive rate of change in the model output. Therefore the negative values in the gradient are discarded to only retain the parts of input that positively contribute to the solution, leading to the following saliency map:
\begin{equation}
L_{Gradient}^c= \|\mbox{ReLU}\left(\frac{\partial y^c}{\partial x}\right)\|\quad,
\end{equation}
where $y^{c}$ is the score for class $c$ before the softmax layer, and $x$ is the input.
While easy to compute and interpret, saliency maps generally perform worse than newer techniques (like CAM, Grad-CAM, and EB), and it was recently argued that saliency maps tend to represent noise rather than the signal of the interest~\cite{kindermans2017reliability}.

Another major limitation of saliency maps is that the class-specific information is not used to create them. However, a good model presumably should be able to use class-specific features to make decisions as the underlying classes can be quite different.
The CAM approach incorporate network activations into spatial localization and provides an improvement over saliency maps   by identifying important, class-specific features at the last convolutional layer as opposed to the input space. It is well-known that such features tend to be more abstract and more semantically meaningful (e.g., faces instead of edges). The downside of CAM is that it requires the layer immediately before the softmax classifier (output layer) to be a convolutional layer followed by a global average pooling (GAP) layer. This precludes the use of more complex, heterogeneous networks, such as those that incorporate several fully connected layers before the softmax layer.

To compute CAM maps, let $F_{k} \in \mathbb{R}^{u \times v}$ be the $k^{th}$ feature map of the convolutional layer preceding the softmax layer. Denote the global average pool (GAP) of $F_{k}$ by
\begin{equation}
e_{k} = \frac{1}{Z}\sum_{i}\sum_{j}F_{k,i,j}
\end{equation}
where $Z=uv$. Then, we can define  class scores, $y^{c}$, as linear combination of the GAP features $e_{k}$:
\begin{equation}
y^{c} = \sum_{k}w^{c}_{k}e_{k},
\end{equation}
where the weights $\w^{c}_{k}$ are learned by training a linear classifier for each class based on the input-output behavior of the network. The weight $w^{c}_{k}$ encodes the importance of feature $k$ for predicting class $c$. By upscaling each feature map to the size of the input images (to undo the effect of pooling layers) the class-specific heat-map in the pixel-space becomes
\begin{equation}\label{CAM}
L^{c}_{CAM}[i,j]=\mbox{ReLU}\left(\sum_{k}w^{c}_{k}F_{k,i,j}\right).
\end{equation}
Zhou et al. show that the heat-map generate localized class-specific feature~\cite{zhou2016learning}.

The Grad-CAM method improves upon CAM by relaxing the architectural restriction that the penultimate layer must be a convolutional. It works for all networks for which the terms $\frac{\partial y^{c}}{\partial F_{k,i,j}}$ are well-defined. In addition, Grad-CAM relaxes the need for training a linear classifier for each class after training the CNN. This is achieved by using feature map weights $\alpha^{c}_{k}$ that are based on back-propagated gradients. Specifically, Grad-CAM defines the weights according to
\begin{equation}
\alpha^{c}_{k} = \frac{1}{Z}\sum_{i}\sum_{j}\frac{\partial y^{c}}{\partial F_{k,i,j}}.
\end{equation}
Following the intuition behind Equation (\ref{CAM}) for CAM, the heat-map in the pixel-space according to Grad-CAM is computed as
\begin{equation}
L^{c}_{Grad-CAM}[i,j]= \mbox{ReLU}\left(\sum_{k}\alpha^{c}_{k}F_{k,i,j} \right),
\end{equation}
where the ReLU function ensures that only features that have a \emph{positive} influence on the class prediction are non-zero.

Excitation Back-Propagation is an intuitively simple, but empirically effective explanation method. The idea is to generate class-specific explanations using class-specific back-propagated error signals.
In~\cite{samek2017evaluating}, it is argued and demonstrated experimentally that explainability approaches such as EB~\cite{zhang2016top}, which ignore nonlinearities in the backward-pass through the network, are able to generate heat-maps that ``conserve'' evidence for or against a network predicting any particular class. Let $a^l_i$ be the i'th neuron in layer $l$ of a neural network and $a^{(l-1)}_{j}$ be a neuron in layer $(l-1)$. Define the \emph{relative} influence of neuron $a^{(l-1)}_{j}$ on the activation $y^l_{i} \in \mathbb{R}$ of neuron $a^l_{i}$, where $y^l_{i}=\sigma(\sum_{ji}W^{l-1}_{ji}y^{(l-1)}_{j})$ and for $W^{(l-1)}$ being the synaptic weights between layers $(l-1)$ and $l$, as a probability distribution $P(a^{(l-1)}_{j})$ over neurons in layer $(l-1)$. This probability distribution can be factored as
\begin{equation}\label{EBP}
P(a^{(l-1)}_{j}) = \sum_{i}P(a^{(l-1)}_{j} \vert a^l_{i})P(a^l_{i}).
\end{equation}
Zhang et al. then define the conditional probability $P(a^{(l-1)}_{j} \vert a^l_{i})$ as
\begin{equation}
P(a^{(l-1)}_{j} \vert a^l_{i}) =
\begin{cases}
Z^{(l-1)}_{i}y^{(l-1)}_{j}W^{(l-1)}_{ji} & \text{if } W^{(l-1)}_{ji} \geq 0, \\
0 & \text{otherwise},
\end{cases}
\label{eq:ebp}
\end{equation}
where  $$Z^{(l-1)}_{i}=\left(\sum_j y^{(l-1)}_{j}W^{(l-1)}_{ji}\right)^{-1}$$ is a normalization factor such that $\sum_{j}P(a^{(l-1)}_{j} \vert a^l_{i}) = 1$. For a given input (e.g., an image), EB generates a heat-map in the pixel-space w.r.t. class $c$ by starting with $P(a^{L}_{i}=c)=1$ at the output layer and applying Equation (\ref{EBP}) recursively.

\section{Explainability for Graph Convolutional Neural Networks}
\label{sec:UDArelated3}
The reviewed explainability methods in the previous section are originally designed for CNNs, which are defined on a signal supported on a uniform grid. We are interested in explainability methods that support non-Euclidean molecular structures, i.e. graphs. In what follows, we first briefly discuss GCNNs and then describe the extensions of these explainability methods to GCNNs.

\subsection{Graph Convolutional Neural Networks}
CNNs have been generalized into GCNNs using two   approaches that extend the notion of convolution to the notion of graph convolution data differently.  Graph  convolution has been defined by incorporating the spatial relation between the nodes~\cite{micheli2009neural,atwood2016diffusion} and using the  spectral  graph  theory~\cite{bruna2013spectral,kipf2017semi}. In this work, we rely on spectral-based definition which is suitable for undirected graphs.

\begin{figure}
\centering
\includegraphics[width=\linewidth]{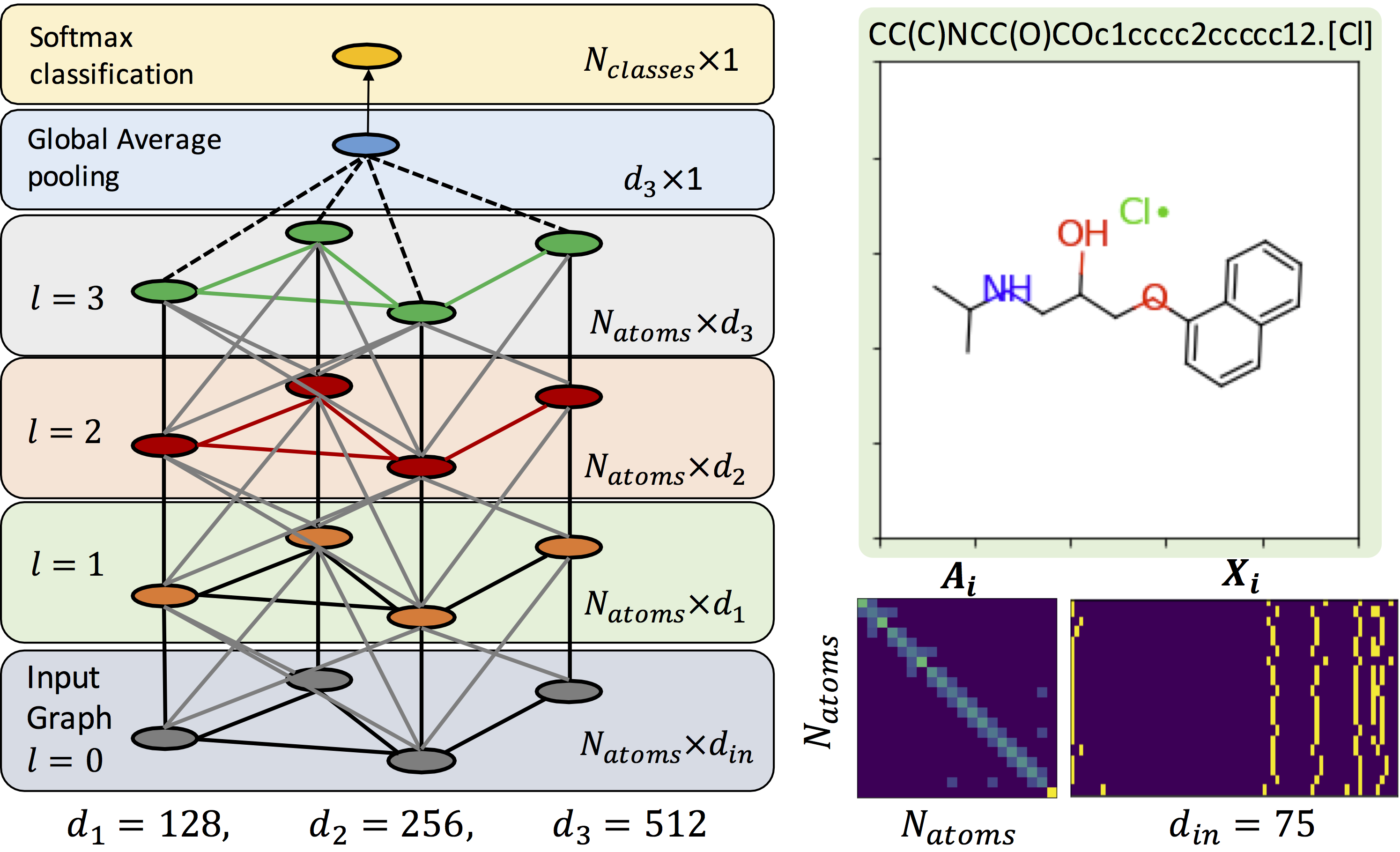}
\caption{Our GCNN architecture together with the visualization of the input feature and adjacency matrix for a sample molecule from BBBP dataset.}
\label{fig:architecture}
\end{figure}

Let an attributed graph with $N$ nodes be defined with its node attributes $X\in\mathbb{R}^{N \times d_{in}}$ and its adjacency matrix $A\in\mathbb{R}^{N\times N}$ (weighted or binary). In addition, let the degree matrix for this graph be $D_{ii}=\sum_j A_{ij}$. Following the work of Kipf and Welling~\cite{kipf2017semi}, we define the graph convolutional layer to be
\begin{equation}
F^{l}(X,A)= \sigma(\underbrace{\tilde{D}^{-\frac{1}{2}} \tilde{A} \tilde{D}^{-\frac{1}{2}}}_{V} F^{(l-1)}(X,A)W^{l})\,,
\label{eq:gcn}
\end{equation}
where $F^{l}$ is the convolutional activations at the $l'th$ layer, $F^{0}=X$, $\tilde{A}=A+I_N$ is the adjacency matrix with added self connections where $I_N\in\mathbb{R}^{N\times N}$ is the identity matrix, $\tilde{D}_{ii}=\sum_j \tilde{A}_{ij}$, $W^{l}\in \mathbb{R}^{d_l\times d_{l+1}}$ are the trainable convolutional weights, and $\sigma(\cdot)$ is the element-wise nonlinear activation function. Figure~\ref{fig:architecture} shows the used GCNN architecture in this work, where the activations in layers $l=1,2,3$ follow Eq. \eqref{eq:gcn}, which is a first-order approximation of localized spectral filters on graphs. Also, note that graph convolution does not distort the graph structure and a graph convolution layer operates on node features. Similar to CNNs, we can come up with  dense layers by computing a global average pooling over node features. By doing so, we convert the graph into a vector which can be considered as a feature extraction method.

For molecule classification, each molecule can be represented as an attributed graph $\mathcal{G}_i=(X_i,A_i)$, where the node features $X_i$  summarize the
 local chemical environment of the atoms in the molecule, including atom-types, hybridization types, and valence structures~\cite{wu2018moleculenet}, and the adjacency matrix encodes atomic bonds and demonstrate the connectivity  of the whole molecule (see Figure~\ref{fig:architecture}). For a given dataset of labeled molecules $\mathcal{D}=\left\{\mathcal{G}_i=(X_i,A_i),y_i\right\}_{i=1}^M$ with labels $y_i$ indicating a certain chemical property, e.g., blood-brain-barrier penetrability or toxicity, the prediction task is to learn a classifier that maps each molecule to its corresponding label, $g:(X_i,A_i)\rightarrow y_i$.
Given that our task is to classify individual graphs (i.e., molecules) with potentially different number of nodes, in our experiments we use several layers of graph convolutional layers followed by a global average pooling (GAP) layer over the graph nodes (i.e. atoms). In this case, all graphs will be represented with a fixed size vector. Finally, the GAP features are fed to a classifier. To enable applicability of CAM~\cite{zhou2016learning}, we simply used a softmax classifier after the GAP layer.
Similar to CNNs, we can train the weights for a GCNN using stochastic gradient descent.

\subsection{Proposed Explainability Methods}
Our goal is to identify the important substructures in the input molecules that contribute do decisions made by the GCNN. The hope is to discover FGs using suitable explainability method.
In this subsection, we describe the extension of CNN explainability methods to GCNNs. Let the $k$'th graph convolutional feature map at layer $l$ be defined as:
\begin{eqnarray}
F^l_{k}(X,A)=\sigma(VF^{(l-1)}(X,A)W^l_k) &
\end{eqnarray}
where  $W^l_k$ denotes the $k'th$ column of matrix $W^l$, and $V=\tilde{D}^{-\frac{1}{2}} \tilde{A} \tilde{D}^{-\frac{1}{2}}$ (see Eq. \eqref{eq:ebp}). In this notation, for the $n$'th atom of the molecule, the $k$'th feature at the $l$'th layer is denoted by $F^l_{k,n}$. Then, similar to CNNs, the GAP feature after the final convolutional layer, $L$, is calculated as
\begin{eqnarray}
e_k=\frac{1}{N}\sum_{n=1}^N F^L_{k,n}(X,A)\quad,
\end{eqnarray}
and the class score can be calculated as:
\begin{eqnarray}
y^{c} = \sum_{k}w^{c}_{k}e_{k}.
\end{eqnarray}
Using these notations, the CNNs explainability methods could be extended to GCNNs as follows:

{\bf Gradient-based} atomic heat-maps for the $n$'th atom can be calculated using automatic differentiation as
\begin{eqnarray}
L_{Gradient}^c[n]=\|\mbox{ReLU}\left(\frac{\partial y^c}{\partial X_n}\right)\|\quad,
\end{eqnarray}
We can visualize heat-maps on the molecular structure to see which atoms contribute to the decision made by the GCNN.

{\bf CAM} atomic heat-maps for the $n$'th atom are calculated as
\begin{eqnarray}
L_{CAM}^c[n]=\mbox{ReLU}(\sum_k w_k^c F^L_{k,n}(X,A)))\,.
\end{eqnarray}

{\bf Grad-CAM}'s class specific weights for class $c$ at layer $l$ and for the $k$'th feature are calculated by
\begin{eqnarray}
\alpha^{l,c}_{k}=\frac{1}{N} \sum_{n=1}^N \frac{\partial y^c}{\partial F^l_{k,n}}\quad,
\end{eqnarray}
and the heat-map for the $n$'th atom calculated from layer $l$ is
\begin{eqnarray}
L_{Grad-CAM}^c[l,n]=\mbox{ReLU}(\sum_k \alpha^{l,c}_{k} F^l_{k,n}(X,A))\,.
\end{eqnarray}
Grad-CAM enables us to generate heat-maps with respect to different layers of the network. In addition, for our model shown in Figure~\ref{fig:architecture}, Grad-CAM's heat-map at the final convolutional layer and CAM's heat-map are equivalent $L_{Grad-CAM}^c[L,n]=L^c_{CAM}[n]$ (See~\cite{selvaraju2017grad} for more details). In this work, we report results for $L_{Grad-CAM}^c[L,n]$ as well as
\begin{eqnarray}
L_{Grad-CAM Avg}^c[n]=\frac{1}{L}\sum_{l=1}^L L_{Grad-CAM}^c[l,n]\,.
\end{eqnarray}

{\bf Excitation Backpropagation}'s heat-map for our model is calculated via backward passes through the softmax classifier, the GAP layer, and several graph convolutional layers. The equations for backward passes through the softmax classifier and the GAP layer are
\begin{eqnarray}
\left\{ \begin{array}{lr}
p(e_k)=\sum_c \frac{e_k ReLU(w_k^c)}{\sum_k e_k ReLU(w_k^c)}p(c) & \text{Softmax}\\ \\
p(F^L_{k,n})=\frac{F^L_{k,n}}{N e^k}p(e^k)& \text{GAP}\,,
\end{array}
\right.
\end{eqnarray}
where $p(c)=1$ for the class of interest and zero otherwise.
The backward passes through the graph convolutional layers, however, are more complicated. For notational simplicity, we decompose a graph convolutional operator into
\begin{equation}
\left\{\begin{array}{l}
\hat{F}_{k,n}^l=\sum_m V_{n,m}F^l_{k,m}\\
\\
F^{(l+1)}_{k',n}= \sigma(\sum_{k'}\hat{F}_{k,n}^{l} W^l_{k,k'})\,,
\end{array}\right.
\end{equation}
where the first equation is a local averaging of atoms (with $V_{n,m}\geq 0$), and the second equation is a fixed perceptron applied to each atom (analogous to one-by-one convolutions in CNNs). The corresponding backward passes for these two functions can be defined as
\begin{equation}
\left\{\begin{array}{l}
p(F^l_{k,n})= \sum_m \frac{V_{n,m}F^l_{k,n}}{\sum_n V_{n,m}F^l_{k,m}}p(\hat{F}^l_{k,m})\\
\\
p(\hat{F}^l_{k,n})= \sum_{k'} \frac{\hat{F}^{l}_{k,n}ReLU(W_{k,k'}^{l})}{\sum_k \hat{F}^{l}_{k,n}ReLU(W_{k,k'}^{l})}p(F_{k',n}^{(l+1)})\,.
\end{array}\right.
\end{equation}

We generate the heat-map over the input layer by recursively backpropagating through the network and averaging the backpropagated probability heat-maps on the input:
\begin{equation}
L^c_{EB}[n]= \frac{1}{d_{in}}\sum_{k=1}^{d_{in}} p(F_{k,n}^0)\quad.
\end{equation}
The contrastive extension of $L^c_{EB}$ follows Eq. (8) in~\cite{zhang2016top}; we call this contrastive variant, c-EB.

Upon applying an explainability method on each molecule, we can determine which single atoms are important and contribute more to the decision made by the GCNN for teh prediction task. However, FGs are atomic substructures that can be modeled as connected sub-graphs in an input molecule graph. If we can identify  repetitive substructures in a dataset of molecules, we can consider them to be potential FG that can cause a particular property. As we see in our experiments, we can perform a substrcuture frequency analysis for this purpose.

\section{Experimental Validation}
 \label{sec:UDArelated4}
 This section describes the experimental setup, results of class-specific explanations, and a substructure frequency analysis identifying relevant FGs for each dataset.

\subsection{Experimental Setup}
We evaluated explanation methods on three binary classification molecular datasets, BBBP, BACE, and task NR-ER from TOX21~\cite{wu2018moleculenet}. Each dataset contains binary classifications of small organic molecules as determined by experiment. The BBBP dataset contains measurements on whether a molecule permeates the human blood brain barrier and is of significant interest to drug design.
The BACE dataset contains measurements on whether a molecule inhibits the  human enzyme $\beta$-secretase. The TOX21 dataset contains measurements of molecules for several toxicity targets. We selected the NR-ER task from this data, which is concerned with activation of the estrogen receptor~\cite{mayr2016deeptox}. These datasets are imbalanced. Class ratios for each dataset are reported in Table~\ref{tab:data}.

In addition, we followed the recommendations in~\cite{wu2018moleculenet}, which is the original paper describing the MoleculeNet dataset, for train/test partitioning. In particular, for BACE and BBBP, the so called ``scaffold'' split is recommended by~\cite{wu2018moleculenet}, which partitions molecules according to their structure, i.e. structurally similar molecules are partitioned in the same split. We emphasize that training the GCNNs and the conventional dataset splits are not the contribution of our paper and we simply follow the standard practice for these datasets.

Using 80:10:10 train/validation/test split, we report ROC-AUC and PR-AUC values of our trained model for each dataset in Table~\ref{tab:ROC}. These results are comparable to those reported in ~\cite{wu2018moleculenet}, and confirm that the model was trained correctly.We we can see, GCNNs are effective tools for prediction tasks within chemistry and biochemsitry.

\begin{table}[t!]
\centering
\begin{tabular}{|c|c|c|}
\hline & Positives & Negatives \\
\hline
BBBP & 1560 & 479 \\
\hline
BACE &  691 & 821\\
\hline
TOX21 & 793 & 5399 \\
\hline
\end{tabular}
\caption{Dataset class breakdown}\label{tab:data}
\end{table}

\begin{table}[t!]
\centering
\begin{tabular}{|c|c|c|c|}
\hline & AUC-ROC & AUC-PR \\
\hline
BBBP & 0.991 / 0.993 / 0.960 & 0.994 / 0.990 / 0.949 \\
\hline
BACE & 0.991 / 0.973 / 0.996 & 0.943 / 0.920 / 0.989 \\
\hline
TOX21 & 0.883 / 0.861 / 0.859 & 0.339 / 0.283 / 0.361 \\
\hline
\end{tabular}
\caption{Evaluation results for splits train/validation/test for each dataset.}\label{tab:ROC}
\end{table}

For all datasets, we used the GCNN + GAP architecture as described in Figure~\ref{fig:architecture} with the following configuration: three graph convolutional layers of size 128, 256, and 512, respectively, followed by a GAP layer, and a softmax classifier. Models were trained for 100 epochs using the ADAM optimizer with learning rate $0.001$, $\beta_1=0.9$, $\beta_2=0.999$. The models were implemented in Keras with Tensorflow backend~\cite{chollet2015}.

\subsection{Class-Specific Explanations}

After training models for each dataset, we apply each explanation method on all samples and to obtain a set of scalars over nodes, i.e. a heatmap.

\begin{figure*}
 \centering
 \includegraphics[width=\linewidth]{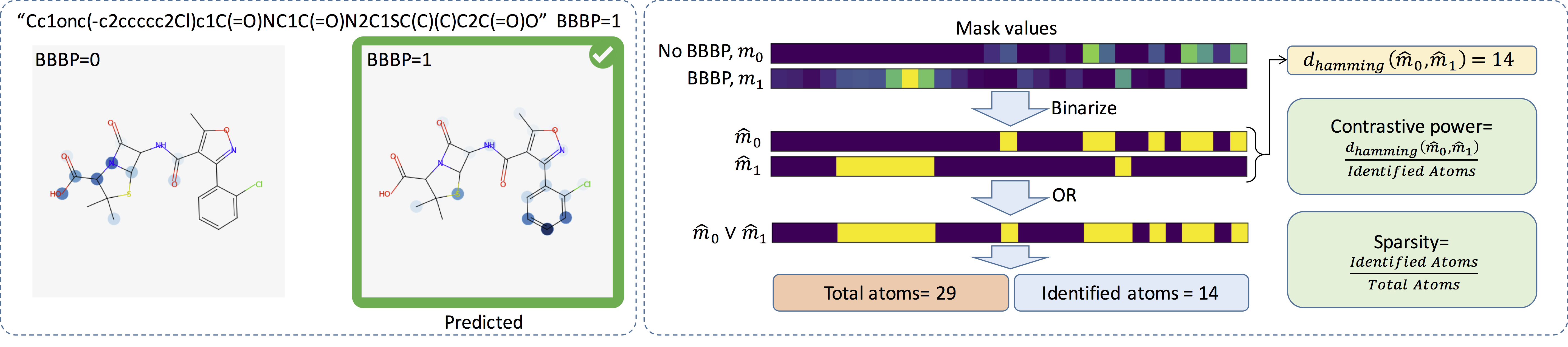}
 \caption{Visualization of the molecule ``Oprea1\_495871'' with molecular formula ``C19H17ClN3O5S\-'', its corresponding SMILES representation, and the result of applying CAM to identify atoms that contribute to its BBBP characteristic  (on the left), and the process of measuring contrastive power and sparsity of the method (i.e., CAM) for this molecule. }\label{fig:contrast}
 \end{figure*}

\begin{figure*}
 \centering
 \includegraphics[width=\linewidth]{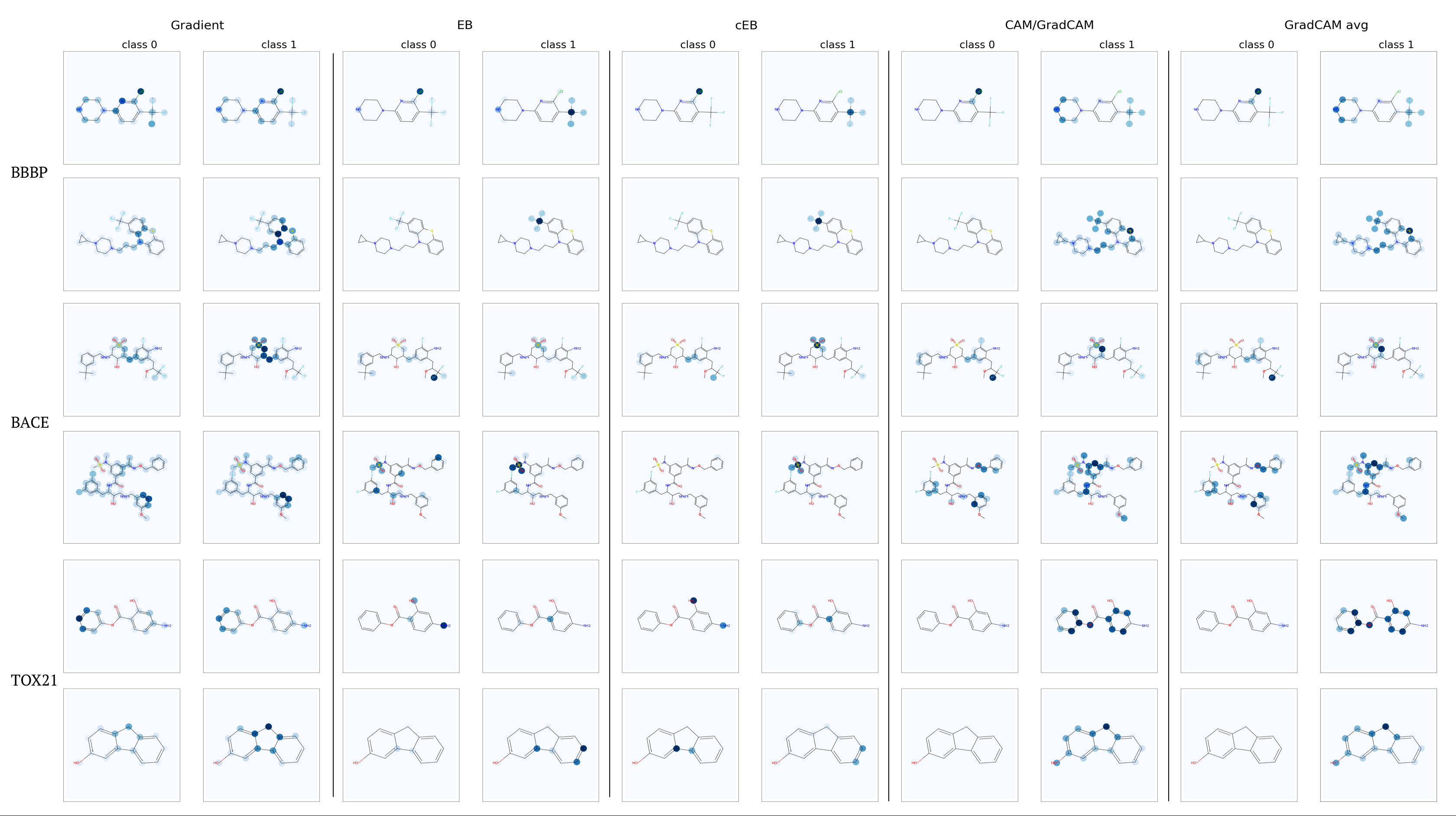}
 \caption{ Selected explanation results for each dataset, e.g., EB highlights CF3 for BBBP. Each sample is a true positive. Class specificity can be seen by comparing regions across classes for each method. A darker blue color indicates a higher relevance for a given class. The results for CAM and Grad-CAM are identical (see Methods). Best viewed on a computer screen.}\label{fig:explain_results}
 \end{figure*}

In Figures~\ref{fig:contrast}, we can see a particular molecule with its corresponding SMILES representation. We have visualized the result of applying CAM to identify atoms that contribute to its BBBP characteristic  (on the left).
 In Figure~\ref{fig:contrast}, scalar importance values are encoded as  the intensity of blue disk over each atom (white: low, blue: high).
We show other selected results  in Figure~\ref{fig:explain_results}.
The heat-maps are calculated for positive and negative classes and normalized for each molecule across both classes and nodes to form a probability distribution over the input nodes. Class specificity can be seen by comparing explanations across classes within a method, i.e., when nodes activated by one class tend to be inactivate for the other.

As opposed to images, where human intuition can help to judge whether the explainability method generates suitable results, it is not as easy to judge which method generates helpful explanations.
For this reason, we define three quantitative metrics that capture desirable aspects of explanations: fidelity, contrastivity, and sparsity. We can then use these metrics to compare the four methods that we have developed.

Fidelity: this metric is calculated to capture the intuition that occlusion of salient features identified through explanations should decrease classification accuracy. More precisely, we define fidelity as the difference in accuracy obtained by occluding all nodes with saliency value greater than $0.01$ (on a scale 0 to 1). We then averaged the fidelity scores across classes for each method. We report these values in Table~\ref{tab:metrics}. The Contrastive Gradient method showed highest fidelity.

Contrastivity: we use this metric  to capture the intuition that class-specific features highlighted by an explanation method should differ between classes. More precisely, we define contrastivity as the ratio of the Hamming distance $d_H$ between binarized heat-maps $\hat{m_0}, \hat{m_1}$ for positive and negative classes, normalized by the total number of atoms identified by either method, $\hat{m_0} \vee \hat{m_1}$, $\frac{d_H(\hat{m_0}, \hat{m_1})}{\hat{m_0} \vee \hat{m_1}}$. We report this metric in Table~\ref{tab:metrics}. Grad-CAM showed the highest contrastivity.

Sparsity: we designed this metric to measure the localization of an explanation. Sparse explanations are particularly useful for studying large graphs, where manual inspection of all nodes is infeasible. More precisely, we define this measure as one minus the number of identified objects in \textit{either} explanation $\hat{m_0} \vee \hat{m_1}$, divided by the total number of nodes in the graph $|V|$, $ 1 - \frac{\hat{m_0} \vee \hat{m_1}}{|V|}$. We report these values in Table~\ref{tab:metrics}. The c-EB method showed the sparsest activations.

% See the Supplementary for a descriptive Figure of con-trastivity and sparsity.

In Figures~\ref{fig:contrast}, we have also visualized the process of measuring contrastive power and sparsity of a method  for a particular molecule (on the right).  Using these metrics, we compare the quality of each explanation method and demonstrate the trade-offs of each method. In short, we conclude three main points:

(1) Grad-CAM is the most contrastive, has the second highest fidelity, but low sparsity. This method is generally suitable, but may be problematic on large graphs.

(2) c-EB is the most sparse but has low fidelity and the second highest contrastivity. Therefore, this method is most suitable for analyzing large graphs at the expense of low fidelity.

(3) Contrastive gradient (CG) has the highest fidelity, but low sparsity and low contrastivity. The lack of contrastivity makes this method unsuitable for class-specific explanations.

\subsection{Substructure Frequency Analysis}
The metrics that we used for comparing the performance of our methods are designed intuitively. However, coming back to the very goal of work, our methods are going to be useful if the identified substructures correspond to real FGs that experimentally have been validated.
For this reason, we need to see if our methods provide such explanations.
Analyzing a collection of molecules with a given property for common substructures is a known technique for discovering relevant functional groups~\cite{CHEN2014280,LOMBARDO201410}. As the number of all possible substructures in a set of molecules is huge, these methods typically restrict analysis to a set of substructures obtained from a fragmentation algorithm. Here, we close the loop for discovering FGs by identifying functional molecular substructures from the machine learned generated heat-maps.

GCNN explanations (i.e., heat-maps) often occur on co-located regions in a molecule and provide a data driven means of generating candidate substructures. Further, we can analyze generated heat-maps for reoccurring patterns which may yield fruitful insights or help select candidate functional substructures.
Connected substructures often manifest in the saliency map of a graph obtained from an explanation method, naturally yielding candidate substructures for further analysis. We describe an automated method for counting the occurrence of salient substructures. In short, for each dataset, we count the frequency of each substructure observed in explanations. Further, we count the overall prevalence of a substructure in a class, which defines a notion of class-specificity.

% devised an automated method for counting the occurrence of substructures identified from the heat-maps generated by the explanation methods. In short, we counted the frequency of each substructure observed in explanations for the three dataset and further analyzed their class specificity.

To identify each substructure, we took the largest connected components consisting of atoms with explanation value greater than some threshold (here, 0), which we call activated atoms, and edges between such atoms. After extracting the activated connected components as identified by the heat-maps, we count their frequency. This analysis requires comparing molecular substructures, a functionality found in open source computational chemistry libraries such as RDKit. We restricted our method to consider only exact substructure comparisons.

To identify substructures, we took the connected components induced by the set of vertices with saliency value greater than some threshold $\tau \in [0,1]$ (here, $\tau=0$). We call these vertices activated. We collect the connected components induced by the activated vertices, and count their frequency. Counting subgraphs requires testing subgraphs for equivalence, the implementations of which are discussed in the following sections. We use GradCAM as the base explanation method.

More formally, let $\mathcal{G} = \{G_i\}_{i=1}^N$ be a collection of graphs with binary labels $\mathcal{Y} = \{y_i\}_{i=1}^N$. For each graph $G_i = (V_i,E_i)$, and for every vertex $v_j \in V_i$, let $a_j \in [0,1]$ be the associated saliency value. We say that a vertex $v_j$ is \textit{activated} if for threshold $\tau$, $a_j \geq \tau$.
The set of activated nodes for graph $G_i$ induces a subgraph $S_i$ of $G_i$, possibly unconnected. Then, we say that each connected component $c_{ij}$ of $S_i$ where $c_{ij}$ has more than one node, is a subgraph identified by the explanation method.

Let the collection of all identified subgraphs in $\mathcal{G}$ by the explanation method be denoted as $\mathcal{S}$. Next, define the counts associated to each identified substructure $s \in \mathcal{S}$ as $N^s_e$. Further, define $N^s_p$ and $N^s_n$ as the number of times a substructure $s$ occurs in the positively labeled data, and the negatively labeled data respectively.

A prevalent substructure in the dataset may artificially show a high prevalence in the generated heat-maps. To account for this potential imbalance, we counted the occurrences of explanation-identified substructures in both positive and negative labeled data in the dataset. We used these counts to normalize the counts obtained from the explanations and construct two ratios:
$$R^s_e = \frac{N^s_e}{N^s_p + N^s_n}$$
and
$$R^s_p = \frac{N^s_p}{N^s_p + N^s_n}$$
where $N^s_e$, $N^s_p$ and $N^s_n$ are the number of times a substructure $s$ occurs in explanations, the positively labeled data, and the negatively labeled data respectively. The ratio $R^s_e$ measures how often a substructure occurs in explanations. The second one measures how often a substructure occurs in positively labeled data, and serves as a baseline for the first. Note that a high $R^s_p$ corresponds to high class specificity for an identified substructure.

These ratios are sensitive to rare substructures. For instance if a substructure occurs only once and occurs in the explanations then it has $R^s_e=1$. To mitigate this sensitivity, we report only substructures that occur more than $10$ times in the dataset.

\begin{figure*}
 \centering
 \includegraphics[width=\linewidth]{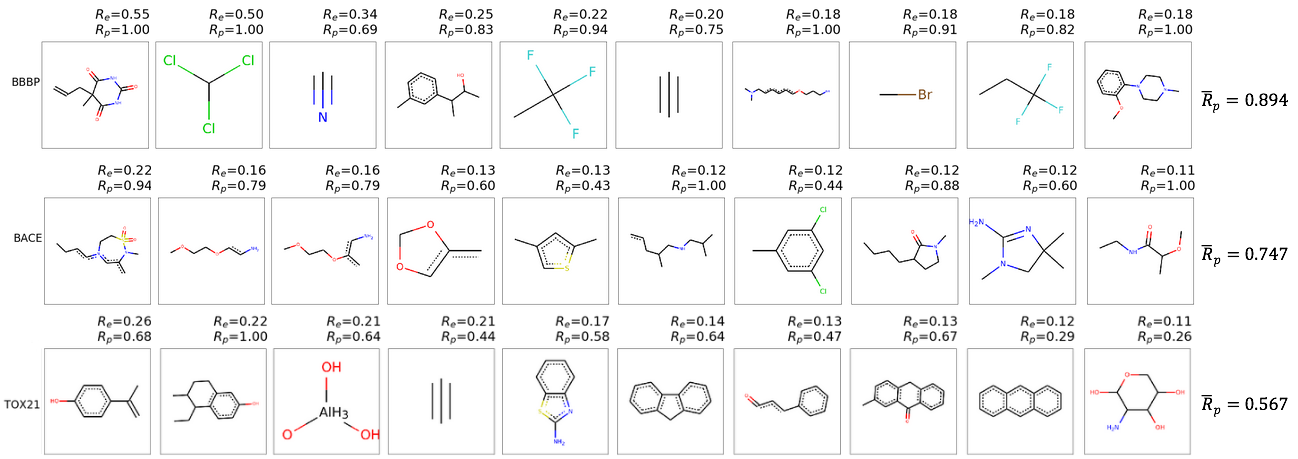}
 \caption{Top 10 most prevalent substructures by dataset. We rank substructures by the ratio $R_e$, the number of times a substructure occurs in explanations over total occurrences in the dataset. For comparison, we also report the ratio $R_p$ of how many times a substructure occurs in the positively labeled set over total occurrences. To account for rare structures, we report only substructures that occurred more than 10 times in the dataset. The right-most column shows average $R_p$ values.}\label{fig:freq_substructs}
 \end{figure*}

Figure~\ref{fig:freq_substructs} shows the most prominent substructures according to our analysis. We have used   Grad-CAM to extract the explanations  because it was the most contrastive method (Table~\ref{tab:metrics}). Additionally, we restricted the explanations to those samples which were true positives. We ranked substructures by $R^s_e$ and report the top 10 for each dataset. As shown in the Figure, the identified substructures have high class specificity and we could identify connected substructures that can be served as candidates for FGs. Interestingly, we observed a few patterns of known FGs being discovered by our method:

1. Halogens (Cl,F,Br) are prevalent in explanations for BBBP~\cite{gentry1999effect}.

2. Amides are prevalent in explanations for BACE~\cite{deutsch1993enzymatic}.

3. Aromatic ring structures are prevalent for TOX21, as validated by prior experiments~\cite{incardona2006developmental}.

Automatic discovery of these FGs validate that we can identify potential FGs using a data-driven scheme without any need to perform experiments. This may be an initial step that can help researchers in the related fields of chemistry and biochemistry to benefit from the power of machine learning.

\begin{table*}[t!]
\centering
{\small
\begin{tabular}{|c||c||c||c||c|}
\hline
&    \textbf{BBBP} & \textbf{BACE} & \textbf{TOX21} & \\
\hline \hline
\multirow{3}{*}{\textbf{CG}}
    &   0.19 & 0.38 & 0.53 & \cellcolor{LightGreen}  \textbf{Fidelity} \\ \cline{2-5}
    &  0.45$\pm$ 2.19& 0.77$\pm$ 2.99 & 0.2 $\pm$ 2.13 & \textbf{Contrastivity} \\ \cline{2-5}
    &  0.22$\pm$ 2.43 & 0.28$\pm$1.58 & 0.21$\pm$2.98 & \textbf{Sparsity} \\
    \hline \hline
\multirow{3}{*}{\textbf{\shortstack{CAM/\\Grad-CAM}}}
    &  0.17 & 0.36 & 0.11 & \textbf{Fidelity} \\ \cline{2-5}
    &   99.99$\pm$ 0.11&  100.0$\pm$ 0.0 &  99.99$\pm$ 0.29 & \cellcolor{LightGreen}  \textbf{Contrastivity} \\ \cline{2-5}
    &  6.26$\pm$7.83& 9.36$\pm$7.67 & 4.86$\pm$8.74 & \textbf{Sparsity} \\
    \hline \hline
\multirow{3}{*}{\textbf{\shortstack{Grad-CAM \\ Avg}}}
    &  0.17 & 0.38 & 0.17 & \textbf{Fidelity} \\ \cline{2-5}
    &   41.06$\pm$19.05& 29.22$\pm$14.04 &44.03$\pm$23.7 & \textbf{Contrastivity} \\ \cline{2-5}
    &  0.01$\pm$0.07& 0.0$\pm$0.0 &0.01$\pm$0.11 & \textbf{Sparsity} \\
    \hline \hline
\multirow{3}{*}{\textbf{EB}}
    &  0.18 & 0.38 & 0.19 & \textbf{Fidelity} \\ \cline{2-5}
    &  50.87$\pm$18.76& 60.29$\pm$15.40 & 49.06$\pm$22.59 & \textbf{Contrastivity} \\ \cline{2-5}
    &  40.35$\pm$22.11 & 51.4$\pm$13.97 & 30.12$\pm$23.04 & \textbf{Sparsity} \\
    \hline \hline
\multirow{3}{*}{\textbf{c-EB}}
    &   0.18 & 0.35 & 0.12 &
    \textbf{Fidelity} \\ \cline{2-5}
    &    96.97$\pm$ 5.68& 97.04$\pm$5.12 & 97.23$\pm$9.3 & \textbf{Contrastivity} \\\cline{2-5}
    &   40.54$\pm$21.69 &  53.01$\pm$13.95 &  31.31$\pm$22.91  & \cellcolor{LightGreen}  \textbf{Sparsity} \\
    \hline
\end{tabular}}
\vspace{-.1in}
\caption{Measures of fidelity, contrastivity, and sparsity for each method. The best performing method (on average) for each metric is highlighted in green (higher values are better).}\label{tab:metrics}
\end{table*}

% \begin{table}[t!]
% \centering
% {\small
% \begin{tabular}{|c|c|c|c|}
% \hline
% & BBBP & BACE & TOX21 \\
% \hline
% CG& 0.45$\pm$ 2.19&  0.77$\pm$ 2.99 & 0.2 $\pm$ 2.13\\
% \hline
% \rowcolor{LightGreen}
% Grad-CAM& 99.99$\pm$ 0.11&  100.0$\pm$ 0.0  & 99.99$\pm$ 0.29\\
% \hline
% {\scriptsize Grad-CAM Avg}& 41.06$\pm$19.05&  29.22$\pm$14.04 &44.03$\pm$23.7 \\
% \hline
% EB& 50.87$\pm$18.76&  60.29$\pm$15.40 & 49.06$\pm$22.59\\
% \hline
% \rowcolor{LightGreen}
% c-EB& 96.97$\pm$ 5.68&  97.04$\pm$5.12 & 97.23$\pm$9.3\\
% \hline
% \end{tabular}}
% \caption{Our measure of contrastive power for different methods. The best performing methods are highlighted in green.}\label{tab:contrast}
% \end{table}

% \begin{table}[t!]
% \centering
% {\small
% \begin{tabular}{|c|c|c|c|}
% \hline
% & BBBP & BACE & TOX21 \\
% \hline
% CG& 99.78$\pm$ 2.43 &  99.72$\pm$1.58 & 99.79$\pm$2.98 \\
% \hline
% Grad-CAM& 93.74$\pm$7.83&  90.64$\pm$7.67  & 95.14$\pm$8.74\\
% \hline
% {\scriptsize Grad-CAM Avg} & 99.99$\pm$0.07&  100.0$\pm$0.0 &99.99$\pm$0.11 \\
% \hline
% EB& 59.65$\pm$22.11 &  48.6$\pm$13.97 & 69.88$\pm$23.04\\
% \hline
% \rowcolor{LightGreen}
% c-EB& 59.46$\pm$21.69 &  46.99$\pm$13.95 & 69.69$\pm$22.91\\
% \hline
% \end{tabular}}
% \caption{Our measure of sparsity of identified atoms,  lower values indicate higher sparsity of the identified atoms. The method with highest sparsity is highlighted in green.}\label{tab:sparse}
% \end{table}

\section{Conclusion}
\label{sec:UDArelated5}
 % 1/4-1/3 column

In this work, we developed a tool that can be used for discovering functional groups for a given class of molecules that are deemed to have similar properties, e.g., toxicity. Our work is based on extending explainability methods, which were designed for CNNs, to GCNNs. We developed four explainability methods and compared these methods qualitatively and quantitatively on three datasets. Our work provides a complementary method that can be used to improve efficiency of the common experimental methods that are used in chemistry and biochemistry.

We compared four methods for identifying molecular substructures that are responsible for a given classification. The GradCAM methods could qualitatively identify functional groups that are known to be chemically active for the specified tasks, e.g., CF3 for brain blood-barrier penetration ~\cite{ghosh2012,kim2016}. This suggests that we can identify some functional groups without preforming experiments. We also identified other  potential functional groups, which can experimentally be tested to confirm their chemical properties.

With our metrics, Grad-CAM and c-EB showed the best contrastive power for showing substructures for different classes. Compared to Grad-CAM, c-EB showed sparser activations, which could be an advantage in certain applications. For the benzene group, however, c-EB could not capture the entire group because the activation was too sparse. Here, GradCAM performed better. So, apparently, there is an optimal value for the sparsity, which may depend on the application.

Our results provide a pathway for automated discovery of relevant functional  groups of molecules. Such an system is a stepping stone towards applications such automated drug discovery. Finally, the proposed framework is a generic tool for discovering functional substructures in general graphs, including social networks, knowledge graphs and electrical grids. As a future application area, we envision that our approach can be used for changing properties of molecules using minimal intervention, e.g., making a toxic molecule non-toxic.

 \bibliographystyle{ieee}
\bibliography{MFGD}


%\vfill


\end{document}